\documentclass[5p,twocolumn]{elsarticle}

\usepackage{multirow,multicol}
\usepackage{tikz}
\usetikzlibrary{arrows}
\usepackage{verbatim}
\usepackage{pgf-pie}
\usepackage{pgfplots}
\usepackage{lineno,hyperref}
\usepackage{subcaption,graphicx}
\usepackage{hyperref}
\usepackage{devanagari}
\usepackage{xcolor}
\usepackage{algorithm,algorithmic}
\usepackage{lscape}
\usepackage{tfrupee}
\modulolinenumbers[5]
\begin{document}

\begin{frontmatter}

\title{Hindi-English Code-Switching Speech Corpus}
\author{Ganji Sreeram, Kunal Dhawan and Rohit Sinha}
\ead{\{s.ganji, k.dhawan, rsinha\}@iitg.ac.in}
\address{Department of Electronics and Electrical Engineering, \\Indian Institute of Technology Guwahati, Guwahati-781039, India.}
\cortext[mycorrespondingauthor]{Corresponding author}

\begin{abstract}
Code-switching refers to the usage of two languages within a sentence or discourse. It is a global phenomenon among multilingual communities and has emerged as an independent area of research. With the increasing demand for the code-switching automatic speech recognition (ASR) systems, the development of a code-switching speech corpus has become highly desirable. However, for training such systems, very limited code-switched resources are available as yet. In this work, we present our first efforts in building a code-switching ASR system in the Indian context. For that purpose, we have created a Hindi-English code-switching speech database. The database not only contains the speech utterances with code-switching properties but also covers the session and the speaker variations like pronunciation, accent, age, gender, etc. This database can be applied in several speech signal processing applications, such as code-switching ASR, language identification, language modeling, speech synthesis etc. This paper mainly presents an analysis of the statistics of the collected code-switching speech corpus. Later, the performance results for the ASR task have been reported for the created database.  
\end{abstract}
\end{frontmatter}
\section{Introduction}
In multilingual communities, the speakers often switch or mix between two or more languages or language varieties during communication in their day to day lives. In linguistics, this phenomenon is referred to as code-switching~\cite{Gumperz_1982_Discourse, nilep2006code}. This phenomenon poses some interesting research challenges to speech recognition~\cite{Lyu_2006_Speech, Bhuvanagirir_2012_Mixed, ahmed2012automatic}, language identification~\cite{Lyu_2008_Language} and language modelling~\cite{Cao_2010_Semantics, Yeh_2010_Integrated, hamed2017building} domains. Over the years, due to urbanization and geographical distribution, people have moved from one place to another for a better livelihood. Hence, communicating in two or more languages helps to interact better with people from different places and cultures. There are many reasons for the occurrence of code-switching. The people belonging to the bilingual communities say that the main reason for code-switching between languages is due to the lack of words in the vocabulary of that particular native language~\cite{Grosjean_1982_Life}. According to ~\cite{myers1993social, malik1994socio, milroy1995one, dey2014hindi}, some possible reasons for code-switching are:  (i) to qualify the message by emphasizing specific words, (ii) to convey a personalized message, (iii) to maintain confidentiality during verbal communication, (iv)  to show expertise, authority, status, etc. Another reason for code-switching is to enrich communication between speakers without any change in the situation. Hence, a native language speaker actively embeds meanings into the conversation by mixing non-native language words~\cite{su2001code}. Based on the locations of the non-native words, code-switching can be broadly classified into two modes. When the switching happens within the sentence it is referred to as the \emph {intra-sentential} code-switching and the one predominantly happening at the sentence boundary is referred to as the \emph{inter-sentential} code-switching~\cite{myers1989codeswitching}. Intra-sentential mode of switching is a common phenomenon and has become an identifying characteristic in bilingual communities. 
     
\begin{figure}[]
     \centering
          \captionof{table}{Example Hinglish sentences showing the inter-sentential code-switching and the variants of the intra-sentential code-switching. Type-1 and Type-2 variants of intra-sentential code-switching refer to high and low contextual information being carried by the non-native (English) words, respectively.}
     \centerline{\includegraphics[width=8.8cm]{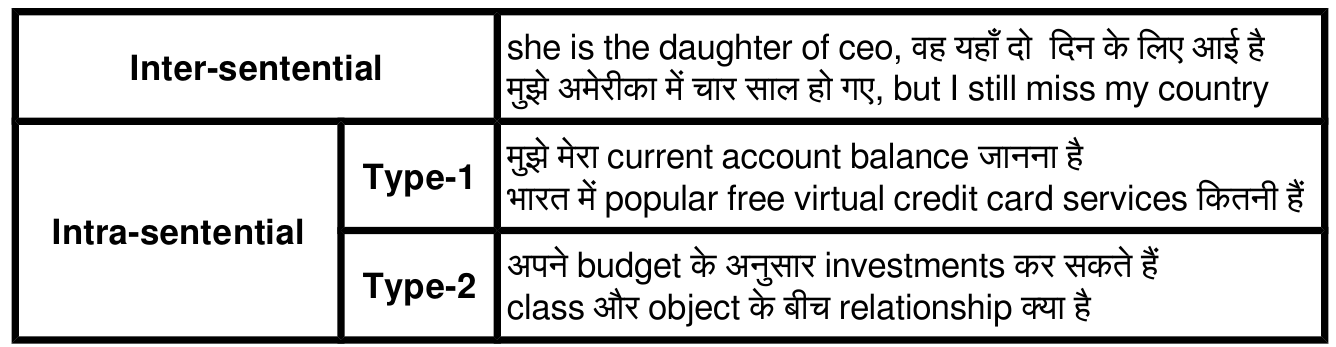}}\vspace{-5mm}
     \label{cs_ex}
  \end{figure}

Over the years, the English language has become the most widely spoken language in the world. After gaining independence from the British rule, though the Indian constitution declared Hindi as the primary official language, the usage of English was continued as a secondary language for its dominance in administration, education, and law~\cite{malhotra1980hindi}. Thereby, the urban population has started a trend to communicate in English for economic and social purposes. Over the years, substantial code-switching to English while speaking Hindi, as well as many other Indian languages, has become a common feature~\cite{kumar1986certain, Smita2009}. Note that, $41.1\%$ of the Indian population are native Hindi speakers and hence the switching between Hindi and English is very common. Also, in the recent past, the researchers have reported that the native language of the speaker influences the foreign (non-native) language acquisition~\cite{flege1995second}. In India, English is taught in schools from elementary level across the country, but very few schools are able to impart correct English pronunciations devoid of native language influences to their pupils. The recent works~\cite{bali2014, Das_2015_Code} have highlighted that the code-switching phenomenon is also observed in chats, comments, and messages posted on the social media sites like Facebook, Twitter, WhatsApp, YouTube, etc. Table~\ref{cs_ex} shows a few example sentences of different modes of code-switching while highlighting the differences in the contextual information carried by the non-native words. In Type-1 intra-sentential code-switching, the non-native language words either occur in sequence or form a phrase, thus carry some contextual information. Whereas, in Type-2 case, the non-native language words are embedded into the native language sentences in such a manner that virtually no contextual information could be derived from those words. Also, during code-switching, we observe that the majority of the sentences belong to Type-2 intra-sentential mode. However, due to lack of availability of the domain-specific resources, the research activity is somewhat limited.

The monolingual automatic speech recognition (ASR) systems may be capable of recognizing a few words from a foreign language but are unable to handle a significant amount of code-switching in the data. On account of the existence of different variants of English pronunciations and code-switching effects, the development of an  ASR system for Hindi-English (Hinglish) code-switching speech data is a challenging task. To the best of our knowledge, there is no large-sized Hinglish corpus available for carrying out the research. Towards addressing that constraint, we recently created a Hinglish corpus covering all typical sources of variations such as accent, session, channel, age, gender, etc. In this work, we describe the details of that corpus and also present basic experimental evaluation is done on the same. 

  The remainder of this paper is organized as follows: In Section~\ref{literature}, we review the code-switching corpora currently reported in the literature. In Section~\ref{sec:database}, the details about Hinglish speech and text corpus along with that of the necessary lexical resources for developing the Hinglish ASR system, are presented in detail. The experimental evaluations using the created Hinglish corpus has been presented in Section~\ref{sec:experiments}. The paper is concluded in Section~\ref{sec:conclusion}.
\section{Literature Review on Code-switching Corpora}
\label{literature}
In literature, a few code-switching speech corpora are already reported and they happen to cover different native and non-native language combinations. In the following, we briefly review those code-switching corpora while summarizing their salient attributes.  
\begin{itemize}
\item The CUMIX Cantonese-English code-switching speech corpus developed by Joyce Y. C. Chan, et al., at the Chinese University of Hong Kong ~\cite{chan2005development}. It contains code-switched speech utterances read by the speakers. The database contains $17$ hours of data read by $40$ speakers. 

\item A small Mandarin-Taiwanese code-switching speech corpus was developed for testing purpose in~\cite{lyu2006speech} by Dau-cheng Lyu and Ren-yuan Lyu. The corpus contains $4000$ Mandarin-Taiwanese code-switching utterances recorded from $16$ speakers. 

\item The English-Spanish code-switching speech corpus was compiled by Franco J. C. and Solorio at the University of Texas~\cite{franco2007baby}. The corpus contains $40$ minutes of transcribed spontaneous conversations of $3$ speakers.

\item The SEAME is a Mandarin-English code-switching conversational speech corpus developed by Dau-Cheng Lyu and Tien Ping Tan from Nanyang Technological University, Singapore, and Universiti Sains Malaysia~\cite{lyu2010seame, vu2012first}. The database contains $63$ hours of spontaneous Mandarin-English code-switching interview and conversational speech uttered by
$157$ Singaporean and Malaysian speakers.

\item Han-Ping Shen, et al., developed the CECOS, a Chinese-English code-switching speech corpus at the National Cheng Kung University in Taiwan~\cite{shen2011cecos}. It contains $12.1$ hours of speech data collected from $77$ speakers uttering prompted code-switching sentences. 

\item A small Hindi-English code-switching speech corpus was collected by Anik Dey and Pascale Fung at Hong Kong University of Science and Technology. This corpus is primarily made up of student interview speech~\cite{dey2014hindi}. It is about $30$ minutes of data collected from $9$ speakers. 

\item A corpus of Sepedi-English code-switching speech corpus was created by the South African CSIR~\cite{modipa2013implications}. The database consists of $10$ hours of prompted speech, sourced from radio broadcasts and read by $20$ Sepedi speakers. 

\item Emre Yılmaz, et al., developed FAME!, a Frisian-Dutch code-switching speech corpus of radio broadcast speech at Radboud University, Nijmegen~\cite{yilmaz2016longitudinal}. The recordings are collected from the archives of Omrop Fryslan, the regional public broadcaster of the province Fryslan. The database covers almost a $50$ years time span.

\item  The  Malay-English corpus developed by Basem H. A. Ahmed, et al., consists of $100$ hours of Malaysian Malay-English code-switching speech data from $120$ Chinese, $72$ Malay and $16$ Indian speakers.~\cite{ahmed2012automatic}. 

\item MediaParl is a Swiss accented bilingual database developed by David Imseng, et al. contains recordings in both French and German as they are spoken in Switzerland. The data was recorded at the Valais Parliament. Valais is a bi-lingual Swiss canton with many local accents and dialects~\cite{imseng2012mediaparl}.

\item The FACST, a French-Arabic speech corpus consists of records of code-switching read and conversational utterances by $20$ bilingual adult speakers who tend to code-switch in their daily lives~\cite{amazouz2018french}. It is about $7.30$ hours of data.

\item A South African speech corpus containing English-isiZulu, English-isiXhosa, English-Setswana, and English-Sesotho code-switching speech utterances is created from South African soap operas by Ewald van der Westhuizen and Thomas Niesler. The soap opera speech is typically fast, spontaneous and may express emotion, with a speech rate higher than prompted speech in the same languages~\cite{van2018first}.

\item The Arabic-English is recently developed by Injy Hamed, et al., by conducting the interviews with $12$ participants~\cite{hamed2018collection}.
\end{itemize}

From the literature review, it can be noted that very small sized code-switching acoustic and linguistic resources have been available so far covering the Indian context. This motivated us to create moderate sized Hinglish resources so that current technological advances in acoustic and language modeling can be explored for Hinglish ASR task.   

\section{Creation of Hinglish Corpus}
\label{sec:database}
This section describes the details of the creation of Hinglish (code-switching) corpus. Firstly, we describe the context and means employed for the creation of Hinglish sentences. Secondly, the details of the procedure followed by the speakers while recording the speech data corresponding to the created Hinglish sentences, are described. Finally, the creation of the lexical resources is discussed.

\subsection{Hinglish text transcripts}
\label{subsec:text_data}
For the experimental purpose, the Hinglish code-switching text data has been collected by crawling a few blogging websites~\footnote{https://shoutmehindi.com}$^{,}$\footnote{https://notesinhinglish.blogspot.in}$^{,}$\footnote{https://www.techyukti.com}$^{,}$\footnote{http://www.learncpp.com} having different contexts. The crawled data is normalized into meaningful sentences and further processed to remove extra spaces, special characters, emoticons, etc. Data thus obtained is used for training the language models, creating the lexicon and also as the text transcription for recording the acoustic data. The salient details of the Hinglish code-switching text corpus created is summarized in Table~\ref{tab:database}

\begin{table}[]
\centering
\caption{The details of the vocabulary size and the word count of the Hinglish code-switching database involved in this study.}
\label{tab:database}
\scalebox{0.9}{
\begin{tabular}{|c|c|c|c|c|}
\hline
\multirow{2}{*}{\textbf{\# sentences}} & \multicolumn{2}{c|}{\textbf{\# words}} & \multicolumn{2}{c|}{\textbf{\# unique  words}} \\ \cline{2-5} 
                                   & Hindi              & English           & Hindi                 & English                \\ \hline
13,071                             & 179,798           & 71,143            & 3,649                 & 4,980                  \\ \hline
\end{tabular}}
\end{table}

\subsection{Hinglish speech corpus}
Hinglish code-switching acoustic data is recorded over the phone from speakers belonging to different states in India.  A consultant was hired for enrolling the speakers to call a toll free number from their mobile phones. The speakers called from various acoustic environments such as home, office, etc. Each speaker was given $100$ unique sentences taken from the above-processed text data. These $100$ sentences are partitioned into $5$ groups which contain $20$ sentences each. Each speaker is requested to record those $5$ groups in $5$ different sessions in order to capture the session variations such as emotions, environment, etc. It is worth highlighting that the duration of the sentences given to each speaker varies from $2$$-$$30$ seconds. Each speaker took about $10$ minutes to complete recording the $20$ sentences in each session. On an average, to complete recording the $100$ sentences, each speaker took about $50$ minutes. The volunteering speakers were compensated with \rupee $250$ for their time and effort. 

The speech data is recorded at $8$ kHz sampling frequency and a bit rate of $128$. This set of speech files was manually inspected and pruned. At the end of the data collection phase, the Hinglish code-switching database contained $7,005$ utterances in total spoken by $71$ speakers.

\subsection{Development of the lexical resources}
For the creation of a lexicon for development of an ASR system for Hinglish data, a unified phone list has been created for Hindi and English words. Also, a unique word list is extracted from the $13,071$ sentences obtained from Sub-section~\ref{subsec:text_data}. 
The phone level transcription for those words has been done manually. Thus created lexicon covers all the pronunciation variations.

\section{Statistical Analysis of the Database}
This section provides the statistical analysis of the  Hinglish code-switching speech corpus. The following subsection provides information about the speakers in the database. Later,  a description of the size and linguistic features of the database is provided.

\subsection{Speaker information}
In order to collect the Hinglish code-switching speech data, the field data consultant recruited speakers from Indian Institute of Technology Guwahati (IITG) who are natively from different states of India.  A total of $71$ speakers are involved in the development of this database. To model a robust ASR system for Hinglish code-switching data, we need to have a database that covers variations due to different geographical distribution, gender, age, etc. Aiming at this, we have collected the database which covers all such variations. The details of the database are discussed below.

\subsubsection{Geographical distribution}
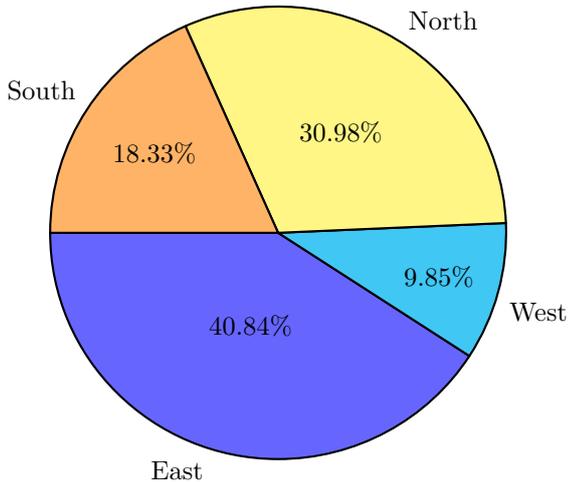
\begin{figure}[h]
\centering
\begin{tikzpicture}
\pie [rotate = 180]
    {
     40.84/East, 9.85/West, 30.98/North, 18.33/South}
     
\end{tikzpicture}
\caption{Geographical distribution of the speakers involved in the collection of database. It is worth highlighting that the collected database covers speakers from $21$ states od India.}\label{fig:loc_dist}
\end{figure}

Since the speakers residing in IITG are from different states of India and from different geographical locations, diversity in the acoustic data is guaranteed. The geographical distribution of the speakers is shown as a pie-chart in Figure~\ref{fig:loc_dist}. The area-wise distribution of the speakers involved in this study is provided in Table~\ref{lab:table1}.

\subsubsection{Age information}
\begin{figure}[h]
\centering
\begin{tikzpicture}
        \begin{axis}[
            symbolic x coords={$<$,20,21,22,23,24,25,26,27,28,29,30,31,32,$>$},
            xtick=data,
                xlabel = Age,
                ylabel = {Population ( $\%$)}
          ]
            \addplot[ybar,fill=blue] coordinates {
                ($<$, 4.22)
                (20, 2.81)
                (21, 8.45)
                (22, 2.81)
                (23, 5.63)
                (24, 5.63)
                (25, 9.85)
                (26, 4.22)
                (27, 18.30)
                (28, 7.04)
                (29, 15.49)
                (30, 5.63) 
                (31, 4.22)
                (32, 1.40)
                ($>$, 2.81)
            };
        \end{axis}
    \end{tikzpicture}\vspace{-4mm}
\caption{Age distribution of the speakers involved in the collection of database. The speakers between $18-62$ years of age are involved in this study.}\label{fig:age_dist}    
\end{figure}
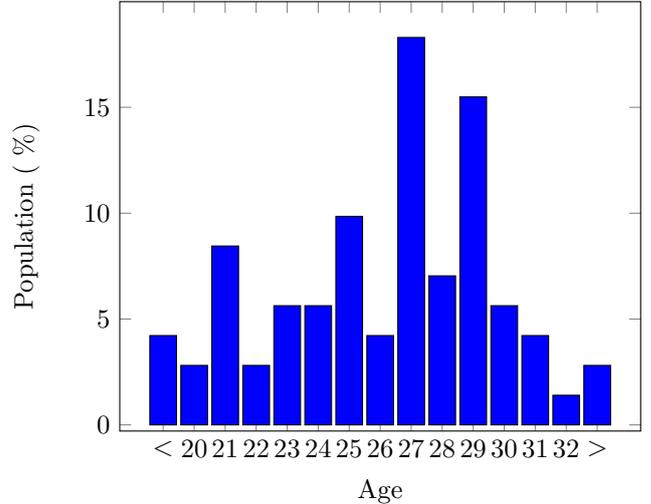

The Hinglish code-switching speech data has been recorded from the speakers between $20$ to  $64$ years of age. The age distribution is shown as a bar-diagram in Figure~\ref{fig:age_dist}.

\subsubsection{Gender information}
\begin{figure}[]
\centering
\begin{tikzpicture}
\pie [rotate = 180]
    {
     61.97/Male, 38.03/Female}
     
\end{tikzpicture}
\caption{Gender distribution of the speakers involved in creation of the Hinglish code-switching database.}\label{fig:gen_dist}
\end{figure}
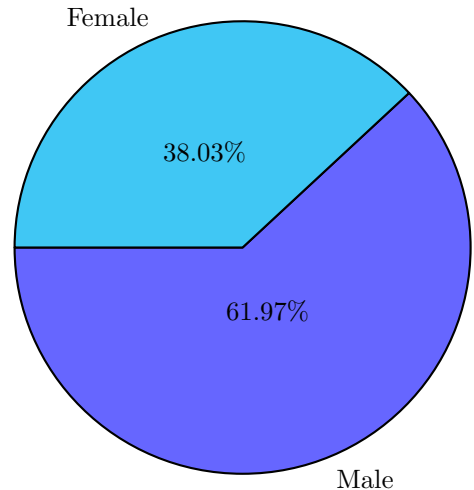

The Hinglish code-switching speech data is recorded from $27$ female speakers and $44$ male speakers resulting in a total of $71$ speakers from different states of India. The gender distribution is shown as a pie-chart in Figure~\ref{fig:gen_dist}.

\begin{table}[]
\centering
\caption{The area and gender wise distribution of the speakers employed for the creation of the Hinglish code-switching database. In total we have $71$ speakers out of which $44$ are male and $27$ are female}
\label{lab:table1}
\scalebox{1}{
\begin{tabular}{|c|c|c|c|}
\hline
\multicolumn{1}{|c|}{\textbf{\begin{tabular}[c]{@{}c@{}}Geographical\\ location\end{tabular}}} & \textbf{\begin{tabular}[c]{@{}c@{}}Male \\ speakers\end{tabular}} & \textbf{\begin{tabular}[c]{@{}c@{}}Female\\ speakers\end{tabular}} & \textbf{\begin{tabular}[c]{@{}c@{}}Total\\ speakers\end{tabular}} \\ \hline 
East                                                                                           & 14                                                                & 15                                                                 & 29                                                                \\ \hline
West                                                                                           & 05                                                                & 02                                                                 & 07                                                                \\ \hline
North                                                                                          & 16                                                                & 06                                                                 & 22                                                                \\ \hline
South                                                                                          & 09                                                                & 04                                                                 & 13                                                                \\ \hline \hline

\textbf{Total}                                                                                          & \textbf{44}                                                       & \textbf{27}                                                        & \textbf{71}                                                       \\ \hline 
\end{tabular}}
\end{table}
\section{Experimental Evaluation and Discussion}
\label{sec:experiments}
The Hinglish code-switching database has been validated by developing an ASR system. For this purpose, the recorded $7005$ utterances are partitioned into training and testing sets containing $5,500$ and $1,505$, respectively. Later, the GMM/DNN based acoustic models are trained using the training set by employing the Kaldi toolkit~\cite{povey2011kaldi}. A $3$-gram language model (LM) is trained over the entire text data obtained from  Sub-section~\ref{subsec:text_data} after excluding those sentences that are used in testing. Therefore, $13071-1505=11,566$ number of sentences are used for training the LM. For developing the $3$-gram LM, we have employed the IRSTLM toolkit~\cite{federico2008irstlm}. The evaluation results in terms of percentage word error rate (\%WER) are given in Table~\ref{eval}. The DNN-based acoustic model with $3$-gram LM resulted in the best $\%WER$ score when compared to other models.
\subsection{Parameter tuning}
The context-dependent GMM acoustic models are trained by tuning the number of senones. After tuning, the number of senones is set to be $2500$. The Gaussian mixtures per senone are set to be $8$ in all the cases. The DNN based acoustic models are trained with $5$ hidden layers and $1024$ nodes with \emph{tanh} as non-linearity function in each of the hidden layers. These models are trained with $20$ epochs and mini-batch size of $128$.
\begin{table}[]
\centering
\caption{Evaluation of Hinglish code-switching speech corpus in context od ASR task. The performance results in terms of percentage word error rate (\%WER) are reported.}\label{eval}
\begin{tabular}{|c|c|c|}
\hline
\textbf{Model} & \textbf{Features} & \textbf{\%WER} \\ \hline \hline
Mono           & MFCC              & 53.51          \\ \hline
Tri1           & MFCC              & 33.52          \\ \hline
Tri2           & MFCC + LDA        & 32.73          \\ \hline
Tri3           & MFCC + LDA + SAT  & 27.20           \\ \hline
DNN            & MFCC + LDA + SAT  & 25.40          \\ \hline
\end{tabular}
\end{table}

\section{Conclusion}
\label{sec:conclusion}  
In this work, the procedure followed to develop a Hinglish code-switching speech database has been presented. It contains $7,005$ utterances spoken by $71$ speakers from different parts of India. The database has been validated by developing an ASR system. The collection of the database is still in progress. 

\section{Acknowledgment}
The authors wish to acknowledge with gratitude for the financial assistance received towards data collection from an ongoing project grant  no. 11(18)/2012-HCC(TDIL) from the Ministry of Electronics and Information Technology, Govt. of India.

\bibliography{refs.bib}

\vspace{0.5cm}

\end{document}